\documentclass[10pt,twocolumn,letterpaper]{article}

\usepackage{iccv}
\usepackage{times}
\usepackage{epsfig}
\usepackage{graphicx}
\usepackage{amsmath}
\usepackage{amssymb}
\usepackage[font={small}]{caption}
\graphicspath{{./figure/}}
\usepackage{algorithm}
\usepackage{algorithmic}
\usepackage{graphicx}
\usepackage{subfig}
\usepackage{color,soul}
\usepackage{array}
\usepackage{multirow}
\usepackage{color}

\usepackage{authblk}
\usepackage[pagebackref=true,breaklinks=true,letterpaper=true,colorlinks,bookmarks=false]{hyperref}
\DeclareMathOperator*{\argmin}{arg\,min}
\iccvfinalcopy 

\def\iccvPaperID{2616} 

\ificcvfinal\fi

\begin{document}

\title{Text Flow: A Unified Text Detection System in Natural Scene Images}

\author[1]{Shangxuan Tian} 
\author[2]{Yifeng Pan}
\author[2]{Chang Huang}
\author[3]{Shijian Lu}
\author[2]{Kai Yu}
\author[1]{Chew Lim Tan}

\affil[1]{ School of Computing, National University of Singapore, Singapore \tabularnewline
\url{tianshangxuan@u.nus.edu, tancl@comp.nus.edu.sg}
}
\affil[2]{ Institute of Deep Learning, Baidu Research, China \tabularnewline
\url{ {panyifeng, huangchang, yukai}@baidu.com } 
}
\affil[3]{ Visual Computing Department, Institute for Infocomm Research, Singapore \tabularnewline
\url{slu@i2r.a-star.edu.sg} 
}

\maketitle

\begin{abstract}
   The prevalent scene text detection approach follows four sequential steps comprising character candidate detection, false character candidate removal, text line extraction, and  text line verification. However, errors occur and accumulate throughout each of these sequential steps which often lead to low detection performance. To address these issues, we propose a unified scene text detection system, namely Text Flow, by utilizing the minimum cost (min-cost) flow network model. With character candidates detected by cascade boosting, the min-cost flow network model integrates the last three sequential steps into a single process which solves the error accumulation problem at both character level and text line level effectively. The proposed technique has been tested on three public datasets, i.e, ICDAR2011 dataset, ICDAR2013 dataset and a multilingual dataset and it outperforms the state-of-the-art methods on \mbox{all} three datasets with much higher recall and F-score. The good performance on the multilingual dataset shows that the proposed technique can be used for the detection of texts in different languages.

\end{abstract}

\section{Introduction}

Machine reading of texts in scene images has attracted increasing interests in recent years, largely due to its important roles in many practical applications such as autonomous navigation, multilingual translation, image retrieval, object recognition, etc. One prevalent scene text detection approach typically consists of four sequential steps namely character candidate detection, false character candidate removal, text line extraction, and \mbox{text} line verification \cite{epshtein2010detecting, huang2013text, yao2014unified, DBLP:dblp_journals/pami/YinYHH14}. However, this prevalent approach suffers from two typical limitations, i.e., the constraint to texts in English and the low detection recall.

\begin{figure}[!t]
	\centering		
			\includegraphics[width=0.32\linewidth, height=0.1\textheight]{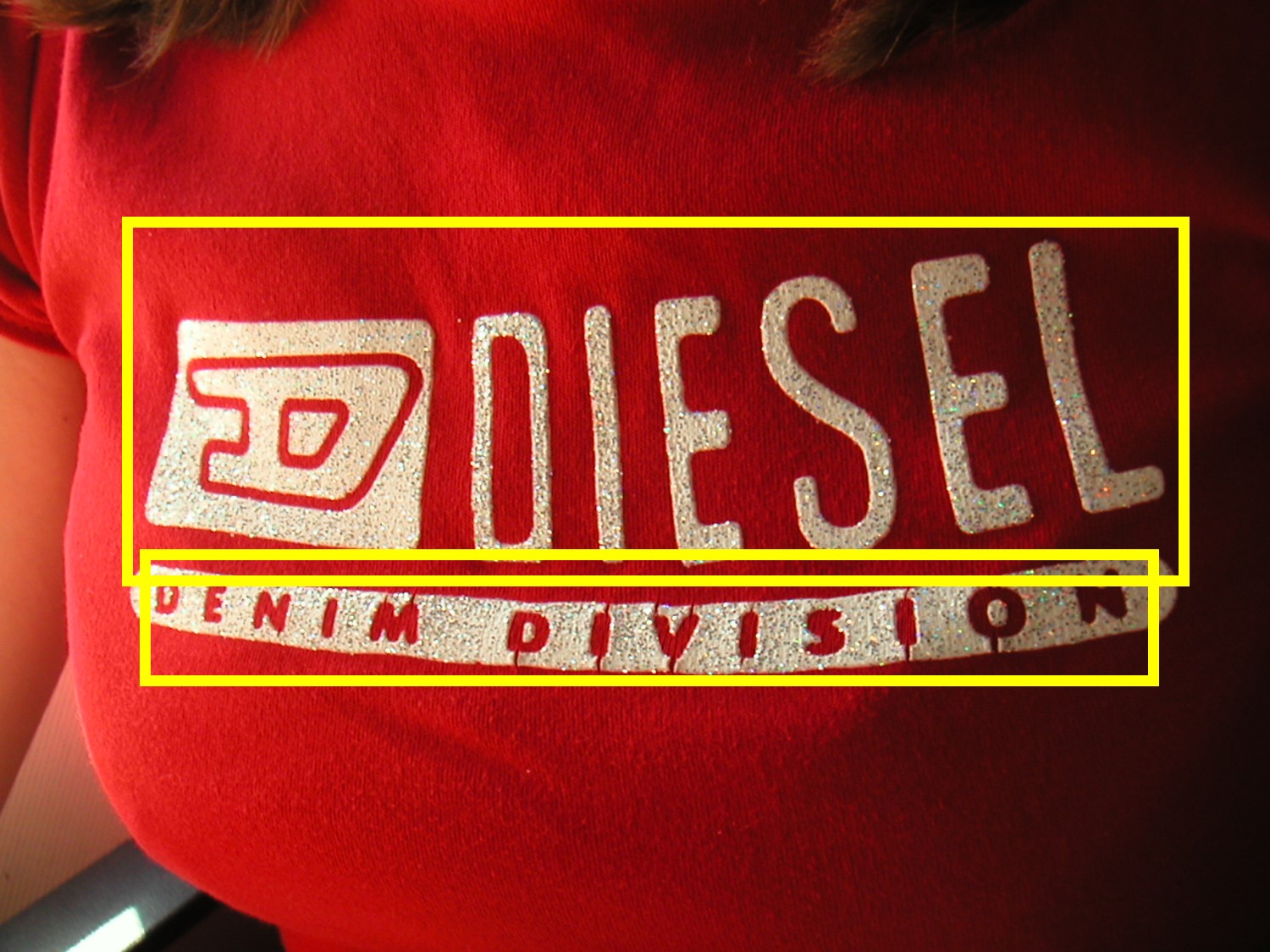}
			\includegraphics[width=0.32\linewidth, height=0.1\textheight]{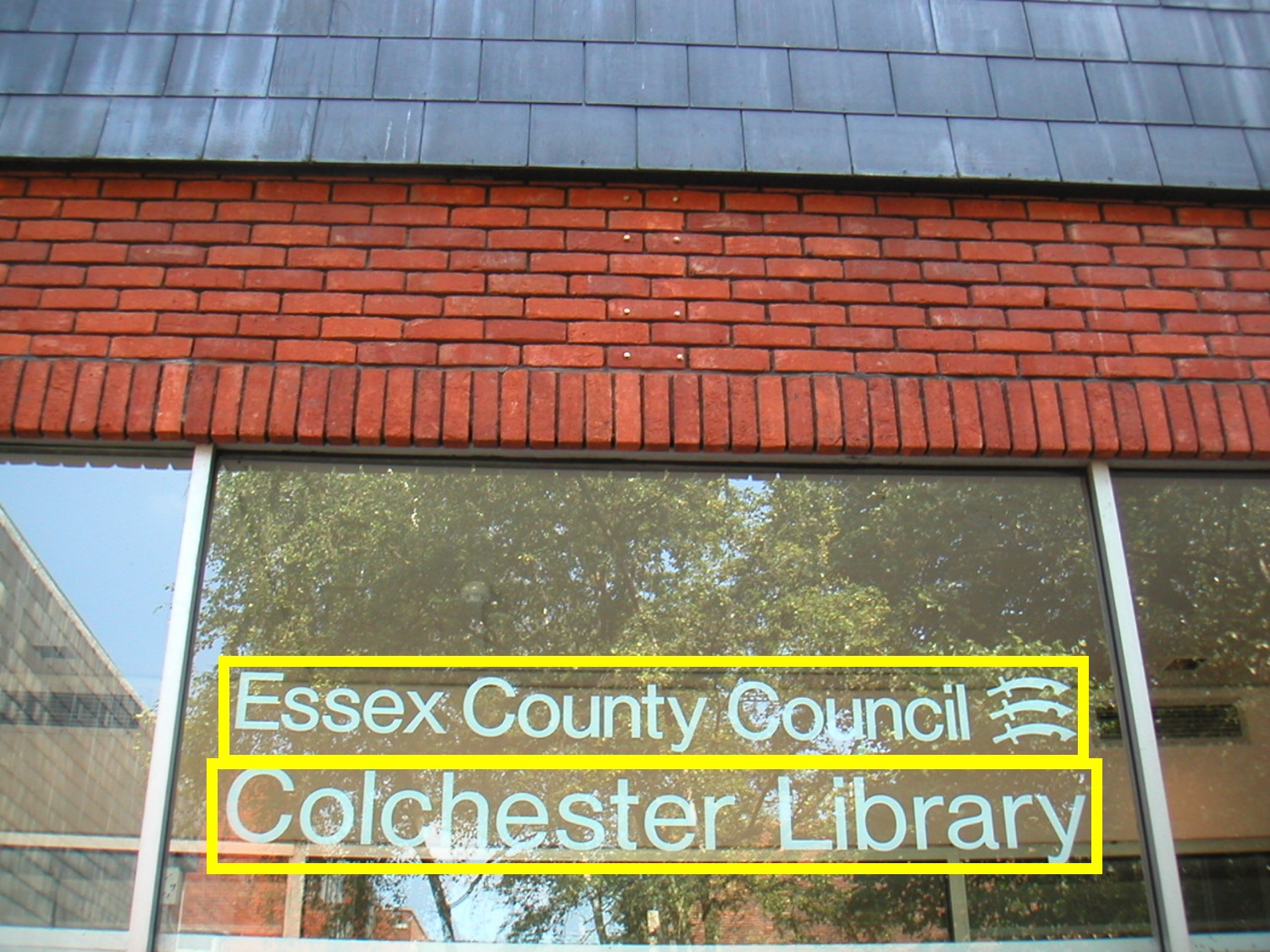}
			\includegraphics[width=0.32\linewidth, height=0.1\textheight]{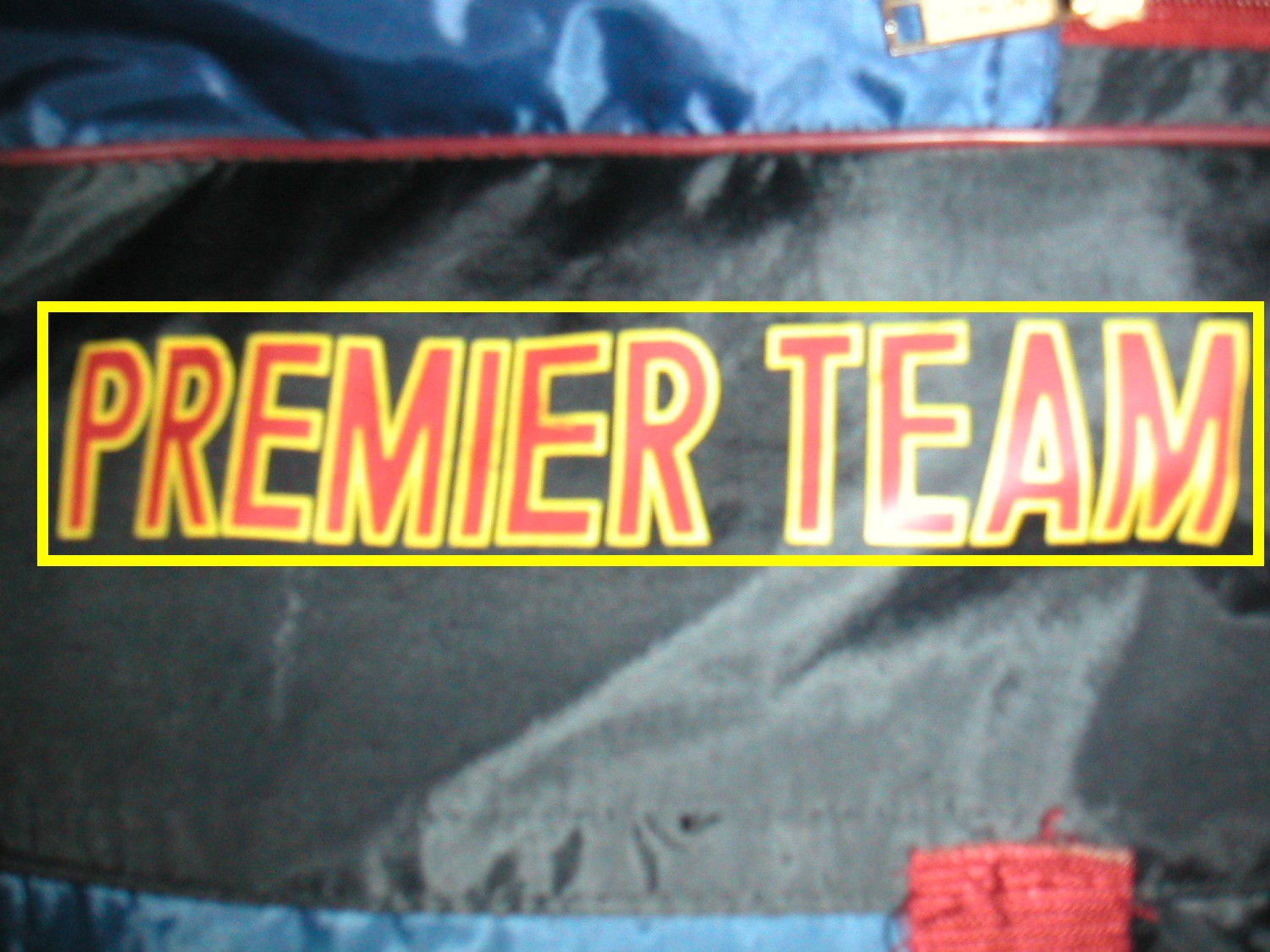} \\ \vspace{1.5mm}
			\includegraphics[width=0.32\linewidth, height=0.1\textheight]{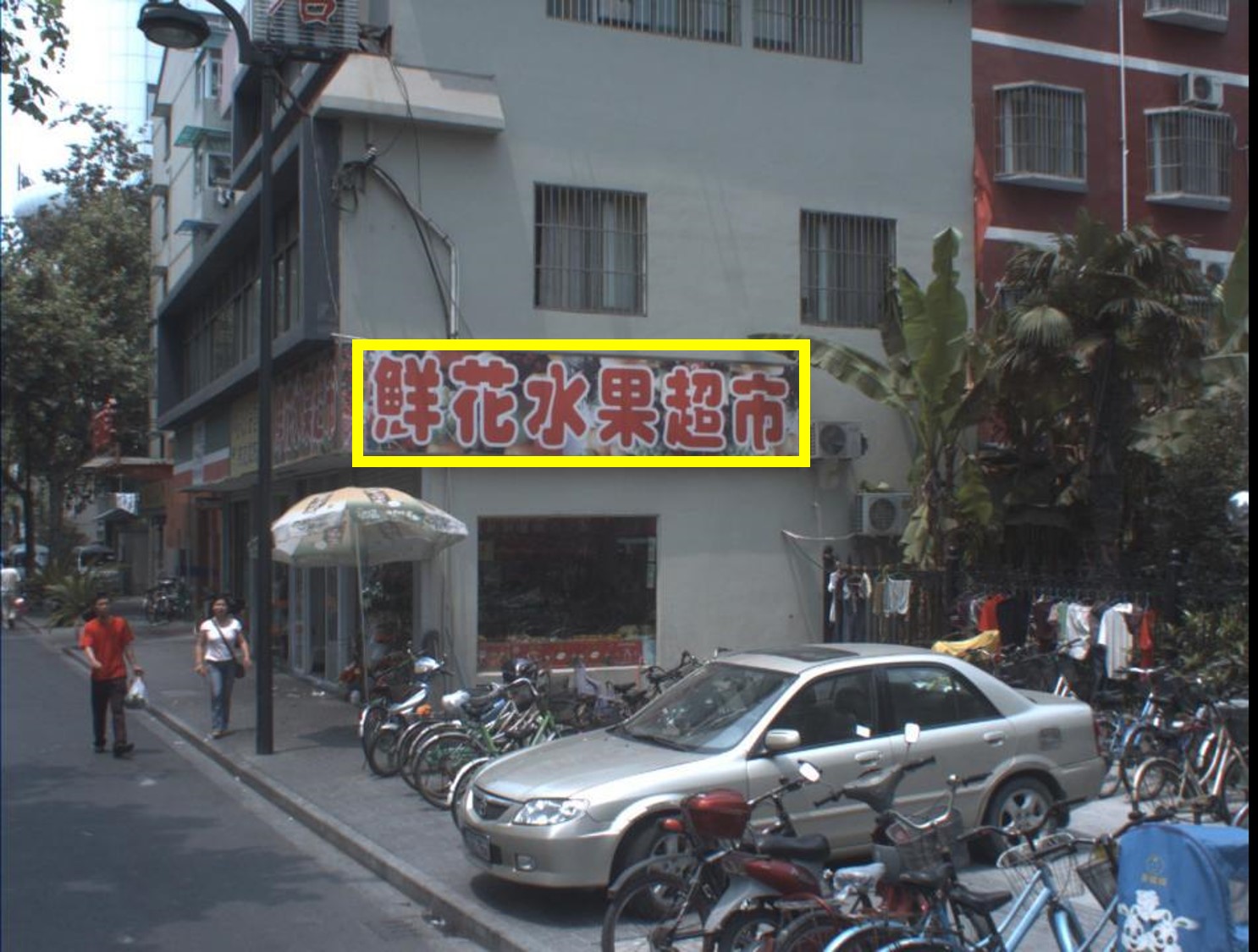}
			\includegraphics[width=0.32\linewidth, height=0.1\textheight]{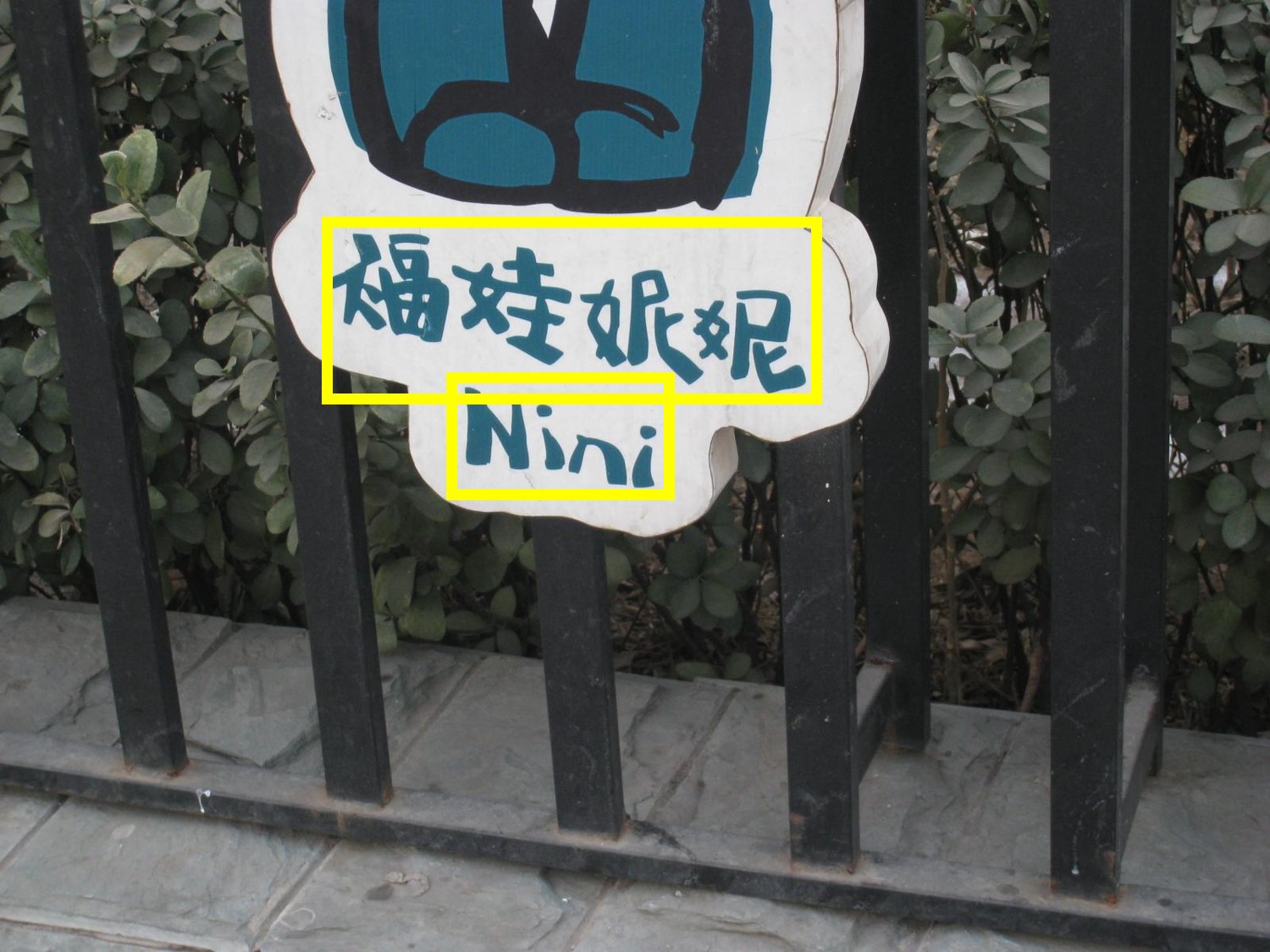}
			\includegraphics[width=0.32\linewidth, height=0.1\textheight]{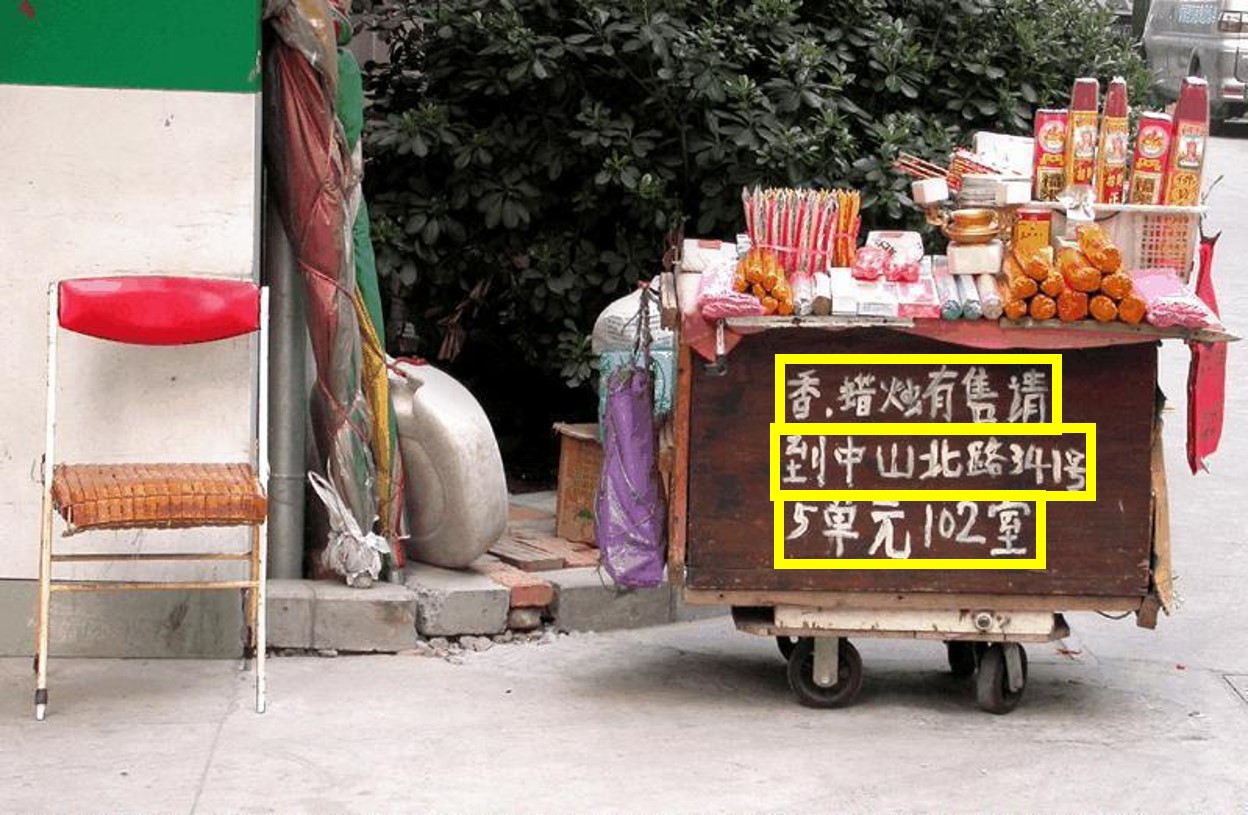}
	\caption{Text detection examples on ICDAR2013 dataset (top row) and the multilingual dataset (bottom row).} 
	\label{Fig:show_result_intro}
\end{figure}
\indent First, character candidate detection often makes use of connected components (CCs) that are extracted in different ways to detect as many text components as possible (for a high recall). On the other hand, this ``greedy'' detection approach includes too many non-text components, leaving the ensuing false alarm removal (for a high precision) a very challenging task. In addition, CCs do not work well for texts of many non-Latin languages such as Chinese and Japanese,  where each character often consists of more than one connected component.

Second, the sequential processing approach often suffers from a typical error accumulation problem. In particular, the error occurring in each of the four sequential steps will propagate to the subsequent steps and eventually lead to a low detection recall. The situation becomes even worse considering that many existing techniques focus on the optimization of simply one or a few of the four sequential steps. In addition, many existing text line extraction techniques rely heavily on knowledge-driven rules \cite{epshtein2010detecting, huang2014robust, neumann2012real} that are unadaptable when conditions change.

We propose a novel scene text detection technique to address these two typical issues with results illustrated in Figure \ref{Fig:show_result_intro}. First, a sliding window based cascade boosting approach is adopted for character candidate detection. One distinctive characteristics of this approach is that character candidates are detected as a whole, hence the complicated process of grouping isolated character strokes into a whole character is not necessary. This feature facilitates the detection of texts of many non-Latin languages such as Chinese where each character often consists of multiple CCs. 

Second, a novel minimum cost (min-cost) flow network model is designed which integrates the last three sequential steps into a single process. The model takes the detected character candidates as inputs and it mainly deals with a unary data cost and a pairwise smoothness cost. The data cost indicates the character confidence and the smoothness cost evaluates the likelihood of two neighboring candidates belonging to the same text line. The problem of text line extraction could hence be formulated into a task of finding the minimum cost text flows in the network.

The min-cost flow model has a number of advantages. First, it extracts text lines with a very high recall since no character-level false alarm reduction is performed before the text line extraction step. Second, it solves the error accumulation problem by combining the character confidence with text layout information, thus eliminates most background noise at both character level and text line level simultaneously. Third, it is simple and easy to implement, as the adopted features are simple and the min-cost flow network problem can be solved efficiently.


The proposed technique has been evaluated on the ICDAR2011 dataset {\cite{shahab2011icdar}}, ICDAR2013 dataset {\cite{karatzas2013icdar}} and a multilingual dataset {\cite{pan2011hybrid}} with texts in both English and Chinese. The experiments show its superior performance and robustness for the detection of texts in different languages.
\section{Related Work}

{The detection of texts in scenes has been studied for years and quite a number of detection systems have been reported. Most of those methods as shown in {\cite{DBLP:dblp_journals/ijdar/LiangDL05, DBLP:journals/pami/YeD15}} typically consist of four sequential steps: text candidate detection, false text candidate removal, text line extraction and text line verification.} 

{Text candidate detection could be roughly grouped into two categories, CCs based methods and sliding window based methods. CCs based techniques detect candidates utilizing bottom up features such as intensity stability and stroke symmetry {\cite{epshtein2010detecting, huang2014robust,  neumann2012real, sun2014robust, DBLP:dblp_journals/pami/YinYHH14}}. However, it is infeasible to process characters composing of multiple components. Sliding window based techniques apply windows of different sizes on the image pyramid and text/non-text classifiers are designed to eliminate noisy windows {\cite{chen2004detecting, jaderberg2014deep, lee2011adaboost, DBLP:dblp_conf/iccv/WangBB11}}. The major limitation is the high computational cost processing numerous windows. However, sliding window techniques have the advantage of incorporating high level texture and shape information, and character with multiple components can be detected as a whole.} 

{Since a large number of non-text candidates are detected in the previous step for a better recall, various text/non-text classifiers such as support vector machine and random forest {\cite{neumann2012real, pan2011hybrid, DBLP:dblp_journals/pami/YinYHH14}}, as well as convolutional neural networks {\cite{bissacco2013photoocr, huang2014robust, Jaderberg_IJCV, sun2014robust}} are adopted to remove false alarms. However, making a hard decision is less reliable when no text line level context is considered.}

{To extract text line from those surviving candidates, a widely adopted technique is hierarchical clustering {\cite{epshtein2010detecting, huang2014robust, DBLP:dblp_journals/pami/YinYHH14}}, which iteratively merges
two candidate text lines if they share a candidate until no text lines could be merged. Graph-based models such as Conditional Random Field (CRF) have been proposed {\cite{pan2011hybrid, shi2013scene}} to label the candidates as text or non-text by graph cut algorithm. Then a learning-based method is presented in {\cite{pan2011hybrid}} which extracts text lines by partitioning a minimum spanning tree into sub-trees. To improve the detection precision, extracted text lines may be further filtered by line level features or average text confidence {\cite{huang2013text, yao2014unified, DBLP:dblp_journals/pami/YinYHH14}}.}

{Existing works {\cite{epshtein2010detecting, huang2013text, yao2014unified, DBLP:dblp_journals/pami/YinYHH14}} focus on increasing the performance of these sequential steps for better detection results. However, the error occurring in the proceeding steps will propagate to the subsequent steps and eventually lead to a low recall. Therefore, an integrated model jointly modeling these sequential steps becomes essential. On the other hand, no integrated model has yet been proposed to solve the text detection problem. There are similar models being utilized in word recognition tasks {\cite{DBLP:dblp_conf/cvpr/MishraAJ12, DBLP:journals/pami/WeinmanLH09}}. In {\cite{DBLP:dblp_conf/cvpr/MishraAJ12}}, a CRF model incorporating unary and pairwise terms is built to model character detections and the interactions between them. The optimal word for the text image is obtained by minimizing the graph energy based on given lexicons.}

{The proposed min-cost flow model is structurally similar to the CRF-based model. However, the CRF model in {\cite{DBLP:dblp_conf/cvpr/MishraAJ12}} is applied on cropped words where layout is simple; the model is unable to determine the corresponding text lines of each character, which needs to be properly addressed in text detection. Besides, the number of detected noisy character candidates is much larger for text detection task. Hence applying the lexicon scheme similar to that in text recognition is less feasible, especially for languages with thousands of character classes such as Chinese. This is due to the fact that text/non-text classifier is more reliable and efficient than character classifier. In addition,  for images with multi-scripts, script needs to be identified first before using the correct lexicon.}  

{Hence, we formulate those isolated steps into an integrated framework, namely Text Flow, where error no longer accumulates and all steps can be jointly optimized in a single model. Meanwhile, false alarms are removed more reliably with line level context. In addition, both Latin and non-Latin scripts are well addressed by the proposed model.}

\begin{figure}[!t]
	\centering		
			\includegraphics[width=1.0\linewidth, height=0.32\linewidth]{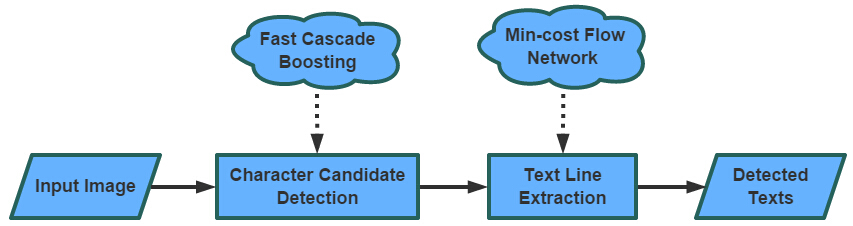}
	\caption{The pipeline of our proposed system.}
	\label{Fig:pipeline}
\end{figure}

\section{Our Proposed System}

Figure \ref{Fig:pipeline} shows the pipeline of the proposed scene text detection system. The character candidate detection is handled by a fast cascade boosting technique. A ``Text Line Extraction'' technique is designed which takes the detected character candidates as inputs and output the verified \mbox{text} lines directly. It integrates the traditional false character candidate removal, text line extraction, and text line verification into a single process and can be solved by a novel mini-cost flow network model efficiently. 

\subsection{Character Candidate Detection}

We detect character candidates by combining the sliding window scheme with a fast cascade boosting algorithm as exploited in \cite{chen2004detecting}. In particular, the cascade boosting in \cite{chen2004detecting} is simplified by ignoring the block patterns in the sliding window. {Furthermore, only six simple features (pixel intensity, horizontal and vertical gradients and second order gradients) are adopted to accelerate the feature extraction process. The features are computed at each pixel location and concatenated as the feature vector for boosting learning.} In fact, fewer feature operations improve the character recall while the weaker character confidence will be later compensated by a convolutional neural network. Positive training examples are the ground truth character bounding boxes and negative examples are obtained by a bootstrap process, hence reducing the chance of each window enclosing multiple characters or single character stroke. 

\begin{figure}[!t]
	\centering		
			\subfloat[]{ \centering \includegraphics[width=0.32\linewidth, height=0.3\linewidth]{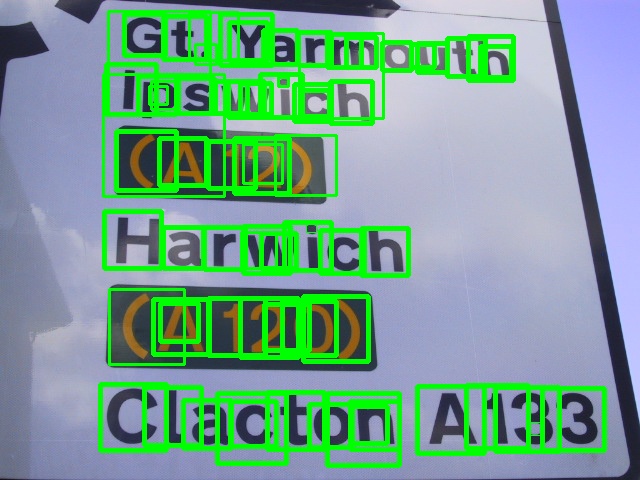}	  	\label{Fig:char_candidates_1}	}  
			\subfloat[]{ \centering \includegraphics[width=0.32\linewidth, height=0.3\linewidth]{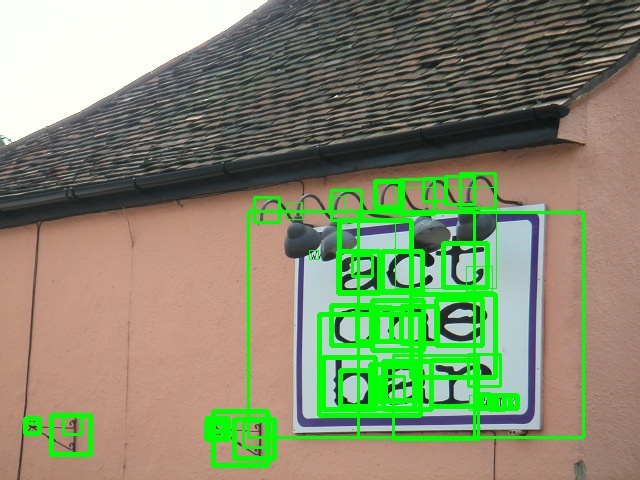}	  	\label{Fig:char_candidates_2}	} 
			\subfloat[]{ \centering \includegraphics[width=0.32\linewidth, height=0.3\linewidth]{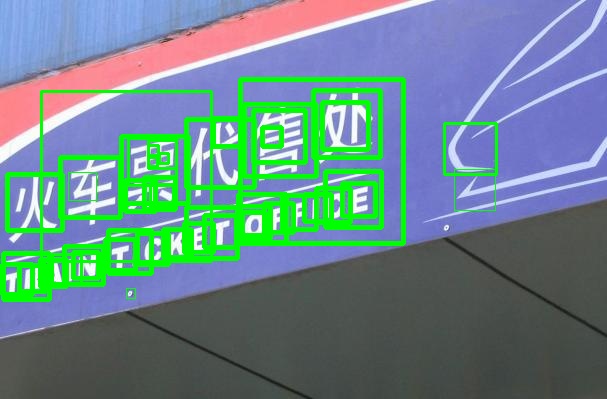}	  	\label{Fig:char_candidates_3}	}
	\caption{Character candidate that are detected by our proposed cascade boosting technique.}
	\label{Fig:char_candidates}
\end{figure}

The sliding window approach is capable of capturing high level text-specific shape information such as the distinct intensity and gradient distribution along the character stroke boundary. In contrast, the CCs based approach focuses on low level features such as intensity stability and is more liable to various false alarms. In addition, the use of the cascade boosting plus some speedup strategies (integral feature map, Streaming SIMD Extensions 2 \cite{guide2010intel}, multi-thread window processing etc.) compensates for the high computational cost of the sliding window process. On the ICDAR2013 dataset, it takes 0.82s per image on average which is comparable to the MSER based technique (0.38s on average \cite{DBLP:dblp_journals/pami/YinYHH14}). Furthermore, the cascade boosting method could detect whole characters instead of isolated components which are taken as negative samples during the training process. This feature helps to reduce the complexity greatly for situations where one single character such as Chinese consisting of multiple isolated components or one CC consisting of several characters (due to the touching). These distinctive characteristics are illustrated in Figure \ref{Fig:char_candidates} where most characters are detected as a whole and few windows contain more than one character.

 {The detected character candidate is considered  positive if $(Area(D) \cap Area(G) )/ (Area(D) \cup Area(G) ) > 0.5$ where $D$ is detected candidate and $G$ is the ground truth character bounding box. Note that each ground truth character window is re-computed as a square bounding box for a fair evaluation. Under this configuration, the proposed approach achieves 23.1\% in precision and 89.2\% in recall.}

\subsection{Text Line Extraction}

We handle the text line extraction by a min-cost flow network model \cite{ahuja1993network} which has been successfully applied for the multi-object tracking problem \cite{zhang2008global}. The target is to integrate multiple scene text detection steps into a single process and accordingly solve the typical error accumulation problem in most existing scene text detection techniques.

A flow network consists of a source, a sink, and a set of nodes that are linked by edges. Each edge has a flow cost and a flow capacity defining the allowed flows across the edge. The min-cost flow problem is to find the minimum cost paths when sending a certain amount of flows from the source to the sink. When applied to the text line extraction problem, the nodes correspond to the detected character candidates and the flows in the network correspond to \mbox{text} lines. We therefore refer to this flow network solution as ``Text Flow''. Intuitively, if we want to extract text flows and meanwhile eliminate non-text candidates both at character level and line level, the network should have a mechanism that deals with three issues: character/non-character confidence, transition constraints and cost between neighboring candidates, and probability of choosing a candidate to be the starting and ending point of a text flow.

\subsubsection{Min-Cost Flow Network Construction}
\label{sec:net_create}
\begin{figure*}[!t]
	\centering		
			\subfloat[]{ \centering \includegraphics[height=0.17\linewidth, width=0.35\linewidth]{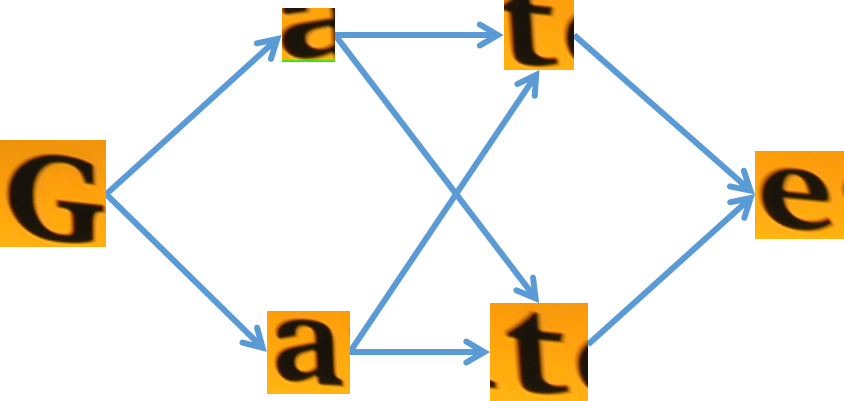}	  	\label{Fig:min_cost_pre}	}  			\hspace{10mm}
			\subfloat[]{ \centering \includegraphics[width=0.55\linewidth]{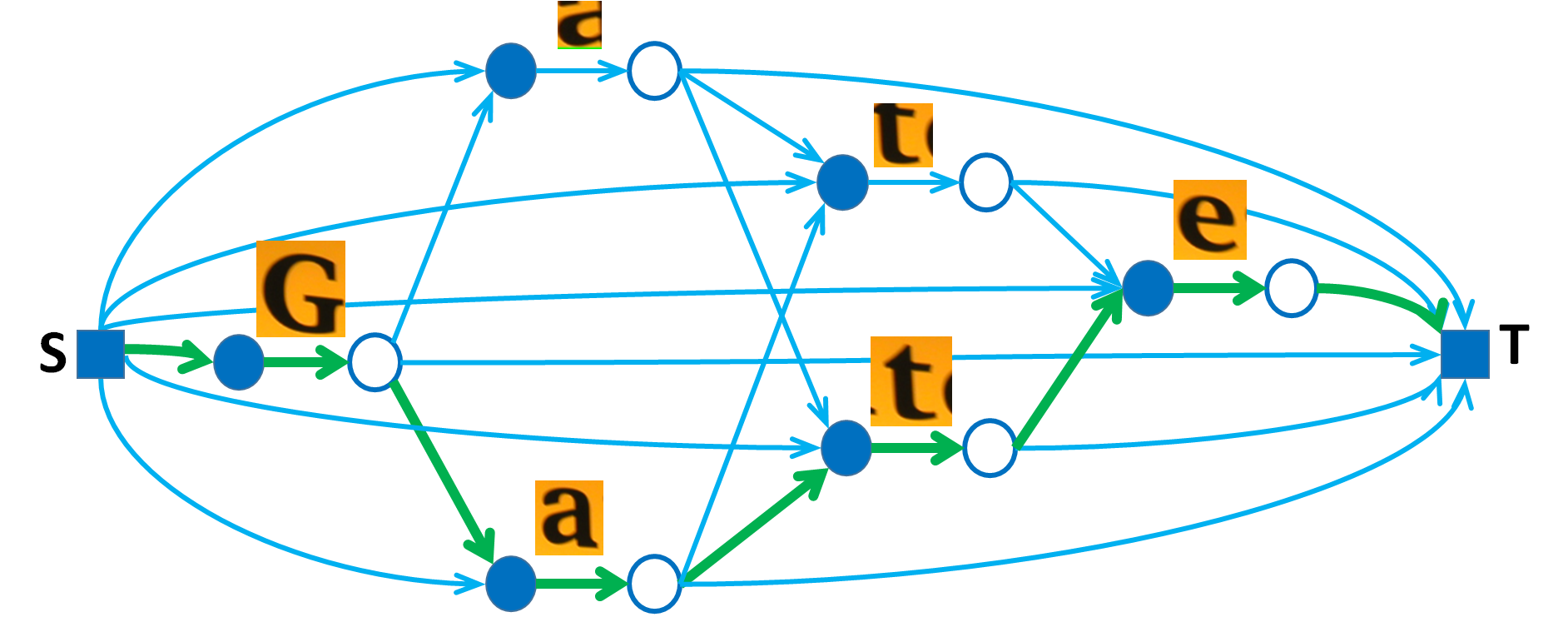}	  	\label{Fig:min_cost_aft}	}  
	\caption{Illustration of the min-cost flow network construction: \protect\subref{Fig:min_cost_pre} shows the six detected character candidates where the edges show the reachability of the detected character candidates. \protect\subref{Fig:min_cost_aft} shows the constructed min-cost flow network. For each candidate in \protect\subref{Fig:min_cost_pre}, a pair of nodes (filled and empty blue circles) are created and an edge linking the two candidates is created and associated with a data cost. A source node ($S$) and a sink node ($T$) (blue rectangles) are created and they are connected to all character candidates in the network. The green path in \protect\subref{Fig:min_cost_aft} shows a true text flow. }
	\label{Fig:min-cost-demo}
\end{figure*}
Based on the assumption that all text lines start from the left to the right, all character candidates are first sorted according to their horizontal coordinates. The flow network can thus be constructed as illustrated in Figure \ref{Fig:min-cost-demo}. First, a pair of nodes are created for each character candidate with an edge in between that represents the \textbf{data cost}. Second, a directed edge from character candidate $A$ to candidate $B$ is created with a \textbf{smoothness cost} if $A$ could reach $B$ based on the transition constraints to be explained later. Third, a source node and a sink node are created and each candidate is connected to both, where the edge connecting with the source has an \textbf{entry cost} and the edge connecting with the sink has an \textbf{exit cost}.
\begin{figure}[!t]
	\centering		
			\includegraphics[width=0.65\linewidth]{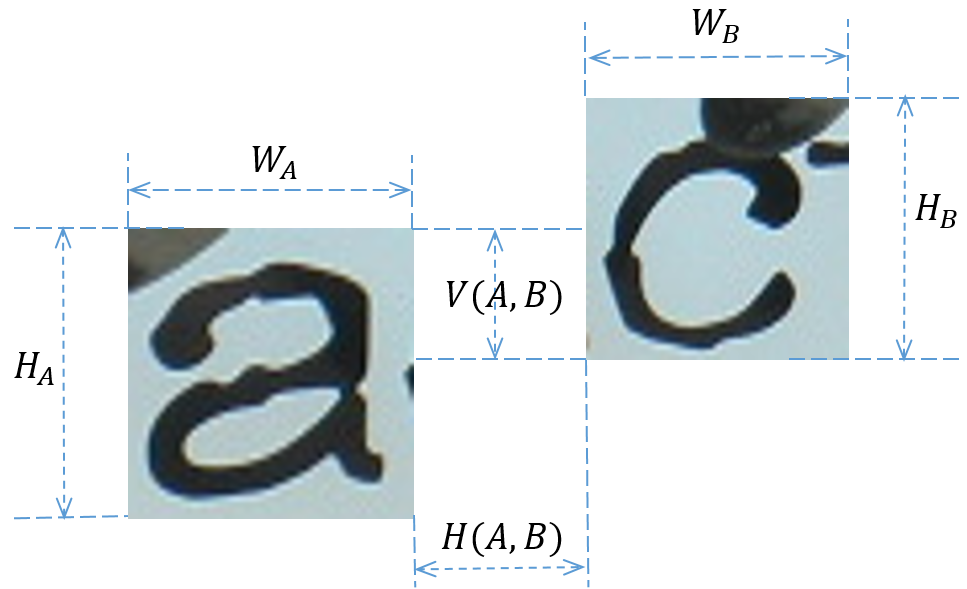}	
	\caption{Spatial and geometrical relationship between two neighboring candidates: Each detected character candidate is represented by a square patch in our system.}
	\label{Fig:dist_def}
\end{figure}
\\
\indent For each character candidate $A$, the next character candidate $B$,  which $A$ could connect to, should be restricted by certain constraints to reduce the errors as well as the search space. Three constrains are employed in our model as illustrated in Figure \ref{Fig:dist_def}. (1) the horizontal distance between $A$ and $B$ should satisfy the condition $H(A, B)/min(W_A, W_B)<T_H$. (2) the vertical distance between $A$ and $B$ should satisfy the condition $V(A, B)/min(W_A, W_B)>T_V$. (3) the size similarity of $A$ and $B$ should satisfy the condition $(|W_A-W_B|)/min(W_A, W_B)<T_S$. Extensive tests on the training datasets show that by setting $T_H, T_V$ and $T_S$ to 2, 0.6, and 0.2 respectively can efficiently reduce the search space yet keep the correct text flows. These conditions can be relaxed to expand the search space in order to detect text lines not complying with the aforementioned constrains.

Figure \ref{Fig:min-cost-demo} shows a simple illustration of the flow network construction process. As Figure \ref{Fig:min_cost_aft} shows, each character candidate is represented by a pair of filled and empty blue nodes with an edge representing the data cost. Likewise, the smoothness cost is associated by an edge between two neighboring character candidates. In addition, each candidate is connected to both the source and the sink with the entry and exit costs, respectively. The text line extraction problem is  to find certain number of text flows that have the minimum cost from the source to the sink. Note that all the costs are represented by edges in network.

\subsubsection{Flow Network Costs}

The costs in the flow network are explained in this part. The data cost $C_1$ is defined as follows: 
\begin{equation}
	\label{eq:data_cost}
	\begin{aligned}
		C_1(A) = - p(Text|A)
	\end{aligned}
\end{equation}
where $A$ is the character candidate image patch and $p(Text|A)$ is the confidence of $A$ being a text region which is measured by a  text/non-text  Convolutional Neural Network (CNN) classifier. The CNN structure is similar to the one that is implemented for the  handwritten digit recognition \cite{lecun1998gradient}. It consists of three convolutionnal layers,  where each layer consists of three steps namely convolution, max pooling and normalization. Two fully connected layers are stacked on the last layer, followed by a softmax layer. 

The CNN is trained by using the image patches that are obtained by applying the character detector on the training images. An image patch (enclosed by a sliding window) is taken as a positive sample if the overlapping with any ground truth patches is larger than 0.5. This step is to minimize the processing error so that the training and testing data are obtained under similar conditions. The data cost is negatively correlated with the confidence of how likely a character candidate is a true character. Higher confidence therefore corresponds to more negative cost which decreases the cost of a text flow passing through it. A true text flow thus will run through character candidates with higher confidence to have a lower cost.


The smoothness cost penalizes two neighboring character candidates that are less likely to belong to the same text line. It exploits two simple features including candidate size and the normalized distance between two candidates. The smoothness cost $C_2$ is defined as follows:
\begin{equation}
	\label{eq:smooth_cost}
	\begin{aligned}
		C_2(A,B) = \alpha*D(A,B) + (1-\alpha)*S(A,B)
	\end{aligned}
\end{equation}
where $D(A,B)$ is the Euclidean distance between the center of character candidates $A$ and $B$ as normalized by the mean of their window widths. $S(A,B)$ is the size \mbox{difference} of $A$ and $B$ defined as $S(A,B) = (|W_A-W_B|)/min(W_A, W_B)$. Parameter $\alpha$ controls the weight of the distance cost and size cost. Note that the smoothness cost is non-negative, i.e., $C_2 \geq 0$. It is large when the two connected character candidates are spatially far away or have very different sizes, meaning that they are less likely to be neighboring characters in a text flow. As a result, a text flow prefers the edge with a smaller smoothness cost while searching for a min-cost flow path.

Though every node has the chance to be the starting/ending of a text line, their probabilities are different. As text lines are usually  linear, intuitively those candidates lying in the middle of a group of candidates are less likely to be the starting/ending point of a text line. The entry cost is therefore defined as follows: 
\begin{equation}
	\label{eq:p_entry}
	\begin{aligned}
		{C_{en}(A)} = -\max_{j}{C_1(j)}
	\end{aligned}
\end{equation}
where $j$ denotes all possible candidates that could reach candidate $A$ in the directed graph. If no candidate reaches $A$, $C_{en}(A)$ is set to $0$. The exit cost can be similarly defined except that $j$ ranges over all the candidates that could be reached by $A$. Equation \ref{eq:p_entry} makes sense because if no character candidate precedes $A$, the chance of a text flow starting at $A$ is large which is consistent with a small entry cost at $A$. On the contrary, the entry cost will increase if there are preceding character candidates in front of $A$. Note that $C_{en}(A)$ not only depends on the spatial position of $A$ but also the text confidence of its preceding candidates. The exit cost  {$C_{ex}(A)$} is defined following the similar idea.

\subsubsection{Min-Cost Flow Network}


To implement the min-cost flow network for text line extraction, the data cost and the smoothness cost (including entry/exit cost) should not be both positive (or negative) to avoid the empty zero-cost flow occasions (or flows having too many candidates). We therefore define the smoothness cost to be positive and the data cost to be negative so that the total cost can be decreased when sending a flow through a series of character candidates. As a result, the min-cost flow prefers a network path consisting of character candidates that have similar size and are close to each other (so as to have smaller positive smoothness costs) and character candidates that have high text confidence (so as to have more negative data cost). A true text flow is highlighted by a  green color path in Figure \ref{Fig:min_cost_aft} because character candidates along this path have much higher text confidence and the neighboring candidates also have similar sizes (the distances between neighboring candidates are roughly the same in this case).


The objective function of the min-cost flow based text line extraction can thus be defined as follows:
{
\begin{equation}
	\label{eq:min_cost_flow}
	\begin{split}
		S = \argmin\limits_{\Gamma} \Big\{ \sum_{i}  C_{en}(i)*f_{en, i} + \sum_{i}\beta*C_1(i)*f_i \\
		+ \sum_{i,j}C_2(i, j)*f_{i,j} + \sum_{i}C_{ex}(i)*f_{i, ex} \Big\} 
	\end{split}
\end{equation}
}
%
where $C_1(i)$ is the unary data cost of a character candidate $i$ and $C_2(i, j)$ is a pairwise cost between two  candidates $i$ and $j$. {$C_{en}(i)$ and $C_{ex}(i)$ are the entry and exit costs of the candidate, respectively}. Parameter $\beta$ is the weight between the data cost and the smoothness cost. {Variables $f_i$, $f_{i, j}$, $f_{en, i}$ and $f_{i, ex}$  represent the number of flows passing through the unary edges, the pairwise edges, and the edges connecting with the source and the sink, respectively.} They should be either 0 or 1 to enforce that each character belongs to at most one text line { and they are determined while solving the min-cost flow problem.} $\Gamma$ denotes all possible flow paths from the source to the sink. The optimal text flows ({which are identified by combinations of $f_i$, $f_{i, j}$, $f_{en, i}$ and $f_{i, ex}$.}) should be those in $\Gamma$  that minimize the overall costs as defined in Equation \ref{eq:min_cost_flow} given the flow number. 


\newcommand*{\factorx}{0.22}
\newcommand*{\factory}{0.3}
\begin{figure*}[!t]
	\centering		
			\subfloat{ 	\centering \includegraphics[width=\factory\linewidth, height=\factorx\linewidth]{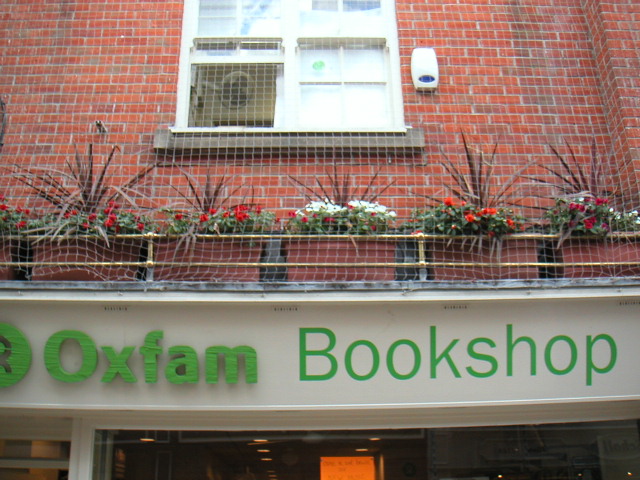}	\enskip
									\centering \includegraphics[width=\factory\linewidth, height=\factorx\linewidth]{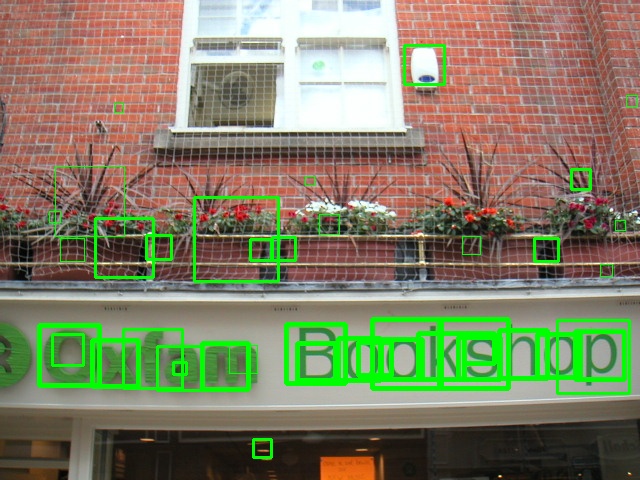}	  	\enskip
									\centering \includegraphics[width=\factory\linewidth, height=\factorx\linewidth]{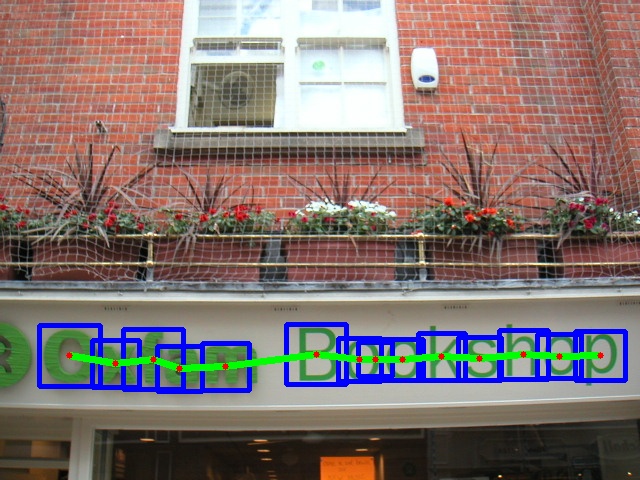}	\label{Fig:text_flow_1}	}  \\
			\subfloat{ \centering \includegraphics[width=\factory\linewidth, height=\factorx\linewidth]{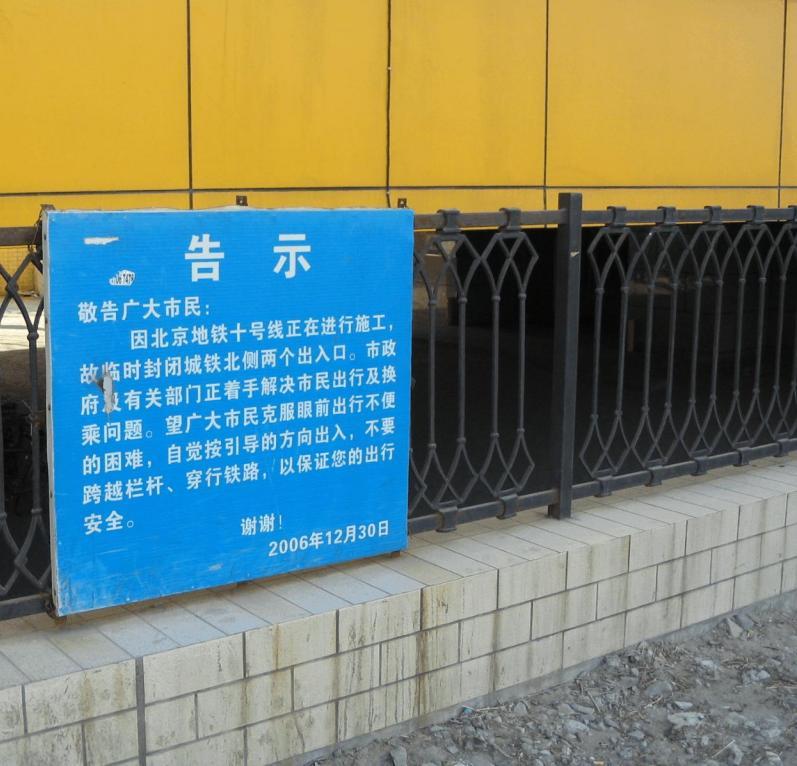}	 \enskip
									\centering \includegraphics[width=\factory\linewidth, height=\factorx\linewidth]{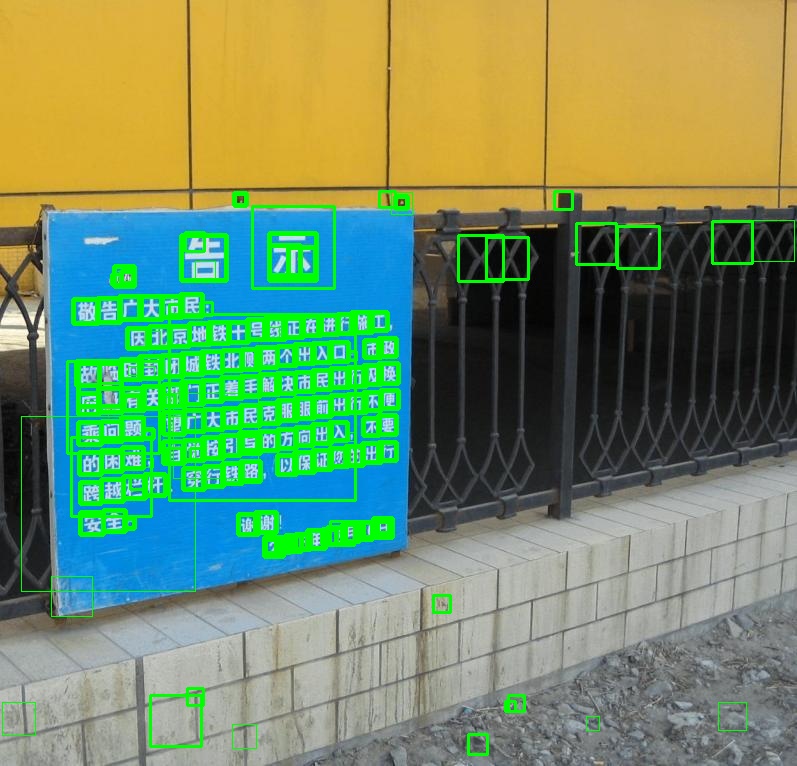} \enskip
									\centering \includegraphics[width=\factory\linewidth, height=\factorx\linewidth]{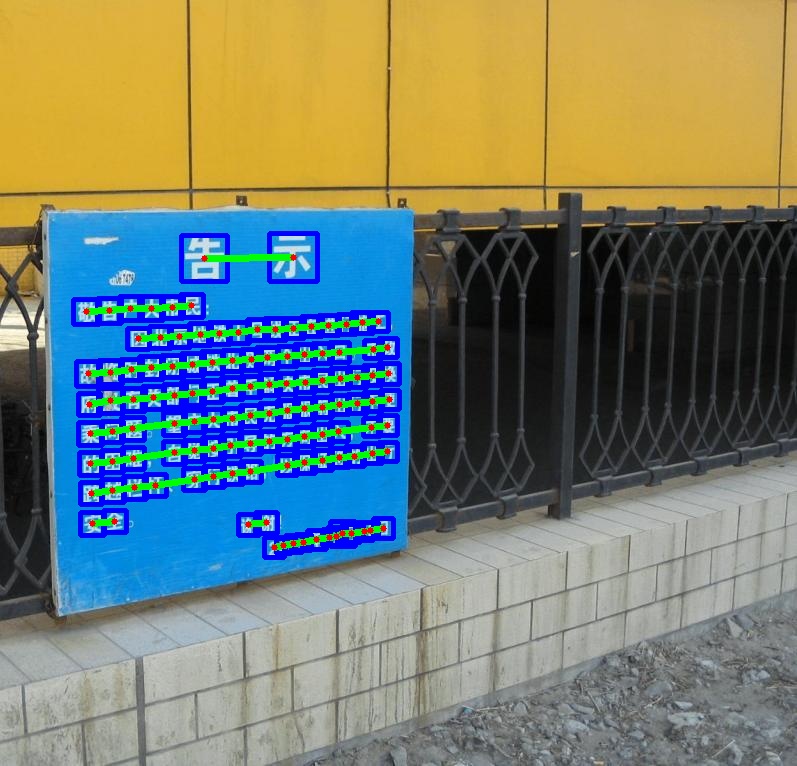}	\label{Fig:text_flow_2}	} 
	\caption{Illustration of text line extraction based on the min-cost flow network model: Input images are shown in the leftmost column and detected character candidates are labelled by green color bounding boxes as shown in the images in the middle column. The extracted text flows are labelled by green lines that link the identified true character (labeled by blue bounding boxes) as shown in the images in the rightmost column. The center of each identified true character is labelled by a red dot.}
	\label{Fig:text_flow}
\end{figure*}

This optimization problem can be efficiently solved by the min-cost flow algorithm \cite{goldberg1997efficient} \footnote{\url{https://github.com/iveney/cs2}}. Algorithm \ref{alg:flow_num} shows how text lines are extracted from the flow network. In particular, one text line is extracted each time as optimization of multiple flows in one go has relatively lower performance. The overlapping characters are removed as described in Step 5, since character candidates are detected at multiple scales and some may overlap with others as illustrated in Figure \ref{Fig:char_candidates}. Algorithm \ref{alg:flow_num} terminates when the flow cost $Cost_i>0$. Our test show that the scene text detection performance is not sensitive to the $\alpha$  in Equation \ref{eq:smooth_cost} and $\beta$ in Equation \ref{eq:min_cost_flow} when both parameters lie in certain ranges ($0.3\leq \alpha \leq 0.5$, $1.5\leq \beta \leq 2.5$). We set $\alpha$ empirically to 0.4 so that the smoothness cost will penalize more on the size difference.
Parameter $\beta$ is set to 2 to make the range of data cost \mbox{two} times of the smoothness cost. This makes the text flow favor more on character candidates with higher text confidence. Under these settings, the cost  will be negative for true text flows and positive otherwise as verified in our experiments. The extracted text flows are shown as green lines running through character candidates in Figure \ref{Fig:text_flow}. As we can see that false candidates are removed during the text line extraction process and text flows do not zigzag  because of the transition constrains as explained in Section \ref{sec:net_create}.
\\
\renewcommand{\algorithmicrequire}{\textbf{Input:}} 
\renewcommand{\algorithmicensure}{\textbf{Output:}}
\begin{algorithm}[!t]
\begin{algorithmic}[1]                 

\REQUIRE Graph $G$ with all the cost precomputed 
\ENSURE  Extracted text flows as well as character candidates in each flow

\REPEAT 
\STATE Set the flow number to 1. 
\STATE Solve the min-cost flow problem by algorithm \cite{goldberg1997efficient} and get the cost for flow $i$ as $Cost_i$.
\STATE Trace the flow path and its character candidates $L_{ij}$ from the algorithm output.
\STATE Delete those character candidates that have more than 50\% overlap with $L_{ij}$ from graph $G$.
\UNTIL{$Cost_i>0$} 

\end{algorithmic}
\caption{Text line extraction by min-cost flow}
\label{alg:flow_num}
\end{algorithm}
\indent The min-cost flow network solution guarantees to find global minimum cost solutions \cite{ahuja1993network, goldberg1997efficient}. Note that beam search {\cite{bissacco2013photoocr}} could also be extended to find the optimal flows in the constructed graph. However, beam search is mostly used to prune the search space, thus it may prune paths that lead to an optimal solution. For our constructed graph, the search space is not large and the min-cost flow network model can produce an optimal solution efficiently.

The extracted text lines can be further split into \mbox{words} for the evaluation on the ICDAR2011 dataset and ICDAR2013 dataset where the ground truth is provided at the word level. We extract words by using the inter-word blank which can be easily detected by projecting the image gradient of each extracted text line to the horizontal axis. The inter-word blank regions usually have very small values along the projected image gradients.

\section{Experiments}

The proposed scene text detection technique has been evaluated on three publicly available datasets, namely, ICDAR2011 {\cite{shahab2011icdar}}, ICDAR2013 \cite{karatzas2013icdar} and a multilingual dataset \cite{pan2011hybrid}. In addition, it has been compared with some state-of-the-art techniques over the three datasets.

\subsection{Data and Evaluation Metric}

The ICDAR2011 dataset consists of 229 training images and 255 testing ones. For each word within each image, the ground truth (for detection) includes a manually labeled bounding box. The ICDAR2013 dataset is a subset of ICDAR2011 dataset since it excludes a small number of duplicated images over training and testing sets and revises ground-truth annotations for several images in the ICDAR2011 dataset. The dataset consists of 462 images including 229 for training and 233 for testing.

The third dataset is a multilingual scene text dataset that was created by Pan \etal as described in \cite{pan2011hybrid}. One motivation of this dataset is for technical benchmarking on texts in non-Latin languages, specifically on Chinese. The dataset consists of 248 images for training and 239 for testing and most images contain texts in both English and Chinese. The ground truth includes a manually labeled bounding box for each text line because texts in Chinese cannot be broken into words without understanding of the text semantics.

The evaluation metrics for these datasets are \mbox{different} as suggested by the dataset creators. For the multilingual dataset, only one-to-one matches are considered and the matching score for a detection rectangle is calculated by the best match with all ground truth rectangles in each image. For the ICDAR2011 and ICDAR2013 datasets, many-to-one (many ground-truth rectangles correspond to one detected rectangle) and one-to-many matches (a ground-truth rectangle corresponds to many detected rectangles) are considered for a better evaluation. The evaluation metrics are described in more details in {\cite{karatzas2013icdar, wolf2006object}}.

\subsection{Experimental Results}


The cascade boosting models for the three public \mbox{datasets} are trained by using the corresponding training images, respectively. The CNN models used in the min-cost flow network are trained using the character candidate samples detected from the training images. In addition, for each character candidate sample in the three public datasets, we create 30 synthetic samples by rotation, shifting, blurring, adding Gaussian noise and so on. The total positive and negative training samples are roughly 600,000 for both.

Tables {\ref{tab:icdar11_res}} and {\ref{tab:icdar13_res}} show experimental results on the ICDAR2011 dataset and ICDAR2013 dataset, respectively. As the two tables show, the proposed technique obtains similar results for the ICDAR2011 dataset and ICDAR2013 dataset and it outperforms state-of-the-art techniques clearly. The winning algorithm in the ICDAR Robust Reading Competition 2013 {\cite{karatzas2013icdar}} reports a F-score of 75.89\% while our text flow technique obtains 80.25\% as shown in Table {\ref{tab:icdar13_res}}. The superior performance can be explained by the proposed min-cost flow model that reduces the error accumulation significantly.

For the multilingual dataset, the first two methods in Table \ref{tab:pan_res} produce state-of-the-art detection performance, where Pan \etal \cite{pan2011hybrid} are actually the creators of the dataset and Yin \etal \cite{DBLP:dblp_journals/pami/YinYHH14} won the Robust Reading Competition 2013. As Table  \ref{tab:pan_res} shows, our proposed technique outperforms the best performing method \cite{DBLP:dblp_journals/pami/YinYHH14} by up to 10\% in detection recall and 7\% in F-score. {To further analyze the performance on the multilingual dataset, we divide the testing dataset into two parts, i.e., Chinese and English, and evaluate the performance separately. There are totally 951 text lines of which 669 are in Chinese ($\sim$70\%) and 282 in English ($\sim$30\%). We manually label the correctly detected text lines as Chinese or English and compute the recall for these two subsets, which are 79.1\% (Chinese) and 76.6\% (English) respectively. Since it is not possible to label a false positive into Chinese or English, the precision cannot be obtained. This result further proves that Text Flow is robust in processing different scripts of languages.}

%
\newcommand{\tabincell}[2]{\begin{tabular}{@{}#1@{}}#2\end{tabular}}
\begin{table}[!t]
\centering
\small
\renewcommand{\arraystretch}{1}
\caption{\small {Text detection results on \textbf{ICDAR2011} dataset (\%)}}
\label{tab:icdar11_res}
\scalebox{0.93} {
		\begin{tabular}{|l|c|c|c|c|}
		\hline
		Method 													& Year & Recall      	&	Precision       	& F-score \\ 		
		\hline
		Kim \etal \cite{shahab2011icdar} & 2011 &62.47 &82.98 &71.28 \\
		\hline
		Huang \etal \cite{huang2013text} & 2013 & 75.00  &82.00 &73.00 \\
		\hline
		Yao \etal \cite{yao2014unified} & 2014 & 65.70  & 82.20  & 73.00 \\
		\hline
		{\textbf{Baseline}}        &-		&{67.13}	
												& {81.48}		&{73.61}  \\
		\hline
		Yin \etal \cite{yinmulti2015} 		& 2015 & 66.01				& 83.77			& 73.84 	\\																						
		\hline
		Neumann and Matas \cite{neumann2013combining}
																		&2013  & 67.50				& 85.40					& 75.40    \\
		\hline
		Yin \etal \cite{DBLP:dblp_journals/pami/YinYHH14}
																		& 2014 & 68.26				& 86.29				& 76.22 	\\
		
		\hline
		Zamberletti \etal \cite{zamberlettitext} & 2014 & 70.00		& 86.00				& 77.00			\\
		\hline
		 Huang \etal \cite{huang2014robust}
																		& 2014 & 71.00				& \textbf{88.00}					& 78.00    \\
		\hline
		Jaderberg \cite{Jaderberg_IJCV} & 2015 & - & - & \textbf{81.00} \\																
		\hline
		\textbf{Text\_Flow} 					&-		& \textbf{76.17}				& 86.24		& 80.89 		\\
		\hline
		\end{tabular}
		}
\end{table}


\begin{table}[!t]
\centering
\small
\renewcommand{\arraystretch}{1}
\vspace{4mm}
\caption{\small {Text detection results on \textbf{ICDAR2013} dataset (\%)} }
\label{tab:icdar13_res}
\scalebox{0.93} {
		\begin{tabular}{|l|c|c|c|c|}
		\hline
		Method 													& Year & Recall      	&	Precision       	& F-score \\ 		
		\hline													
		Shi \etal \cite{shi2013scene}
																		& 2013 & 62.85				& 84.70					& 72.16    \\
		\hline
		\textbf{Baseline} 							&- &66.54				& 80.69					& 72.93 		\\ 									
		\hline
		Ye and David \cite{ye2014scene} & 2014 & 62.26		& 89.17				& 73.33			\\
		\hline
		Yin \etal \cite{yinmulti2015} 		& 2015 & 65.11				& 83.98			& 73.35 	\\																						
		\hline
		Neumann and Matas \cite{neumann2012real}
																		& 2013 & 64.84				& 87.51						&74.49			\\
		\hline
		Yin \etal \cite{DBLP:dblp_journals/pami/YinYHH14}
																		& 2013 & 66.45				& 88.47						& 75.89 	\\
		\hline
		Lu \etal \cite{lu2015scene} & 2015 & 69.58				& \textbf{89.22}				& 78.19    \\
		\hline
		\textbf{Text\_Flow} 				&-	& \textbf{75.89}				& 85.15					& \textbf{80.25} 		\\
		\hline
		\end{tabular}
		}
\end{table}

\begin{table}[!t]
\centering
\small
\renewcommand{\arraystretch}{1}
\vspace{4mm}
\caption{\small {Text detection results on \textbf{Multilingual} dataset (\%)}} 
\label{tab:pan_res}
\scalebox{0.93} {
		\newcolumntype{H}{>{\centering\arraybackslash}p{1.18cm}}
		\begin{tabular}{|l|H|H|H|H|}
		\hline
		Method 													& Recall      	&	Precision       	& F-score 		& Speed (s)\\ 															
		\hline
		Pan \etal \cite{pan2011hybrid}
																		& 65.9				& 64.5						& 65.5			&	3.11	\\
		\hline
		{\textbf{Baseline}}        	&{67.2}	 &{78.6}
				&{72.4}  			&   {0.88} \\
		\hline
		Yin \etal \cite{DBLP:dblp_journals/pami/YinYHH14}
																		& 68.5				& 82.6						& 74.6 			& \textbf{0.22}	\\
		\hline
		\textbf{Text\_Flow} 		& \textbf{78.4}		& \textbf{84.7}		& \textbf{81.4}		&	0.94		\\
		\hline
		\end{tabular}
}
\end{table}

\newcommand*{\factorz}{0.126}
\begin{figure*}[!t]
	\centering		
			\includegraphics[width=0.18\linewidth, height=\factorz\textheight]{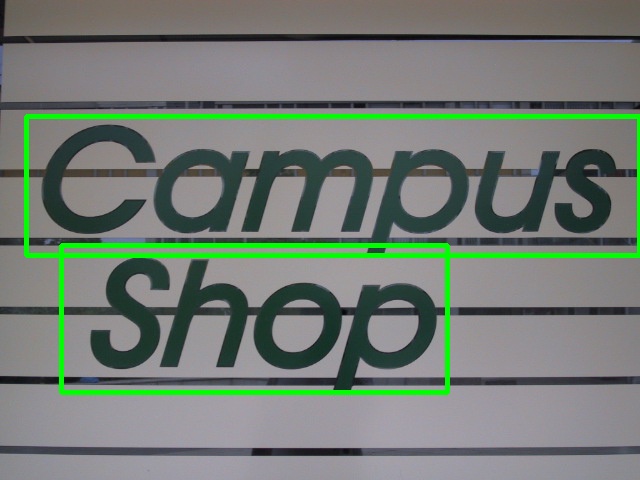} 
			\includegraphics[width=0.18\linewidth, height=\factorz\textheight]{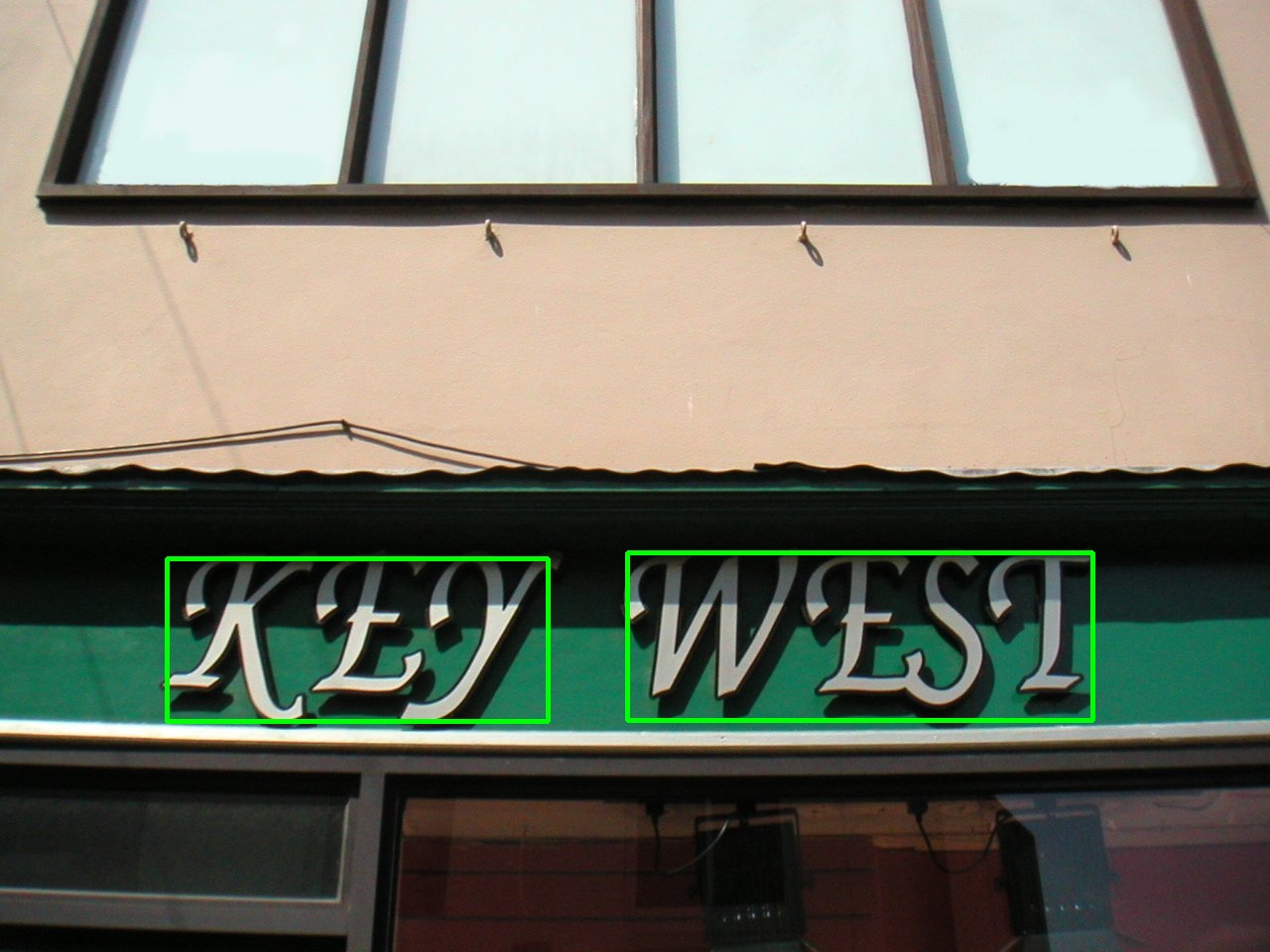} 
			\includegraphics[width=0.18\linewidth, height=\factorz\textheight]{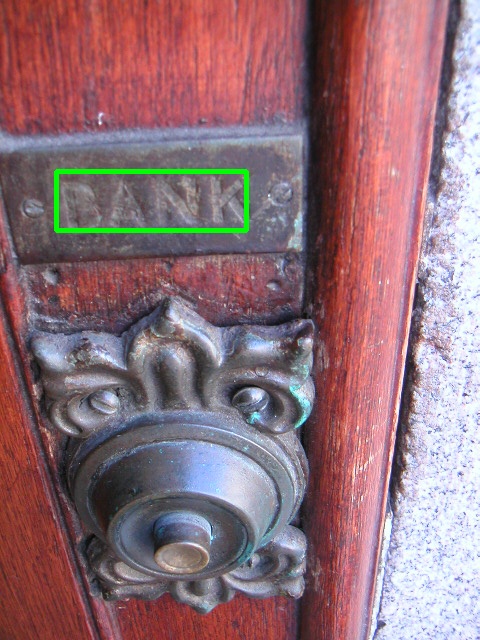} 
			\includegraphics[width=0.18\linewidth, height=\factorz\textheight]{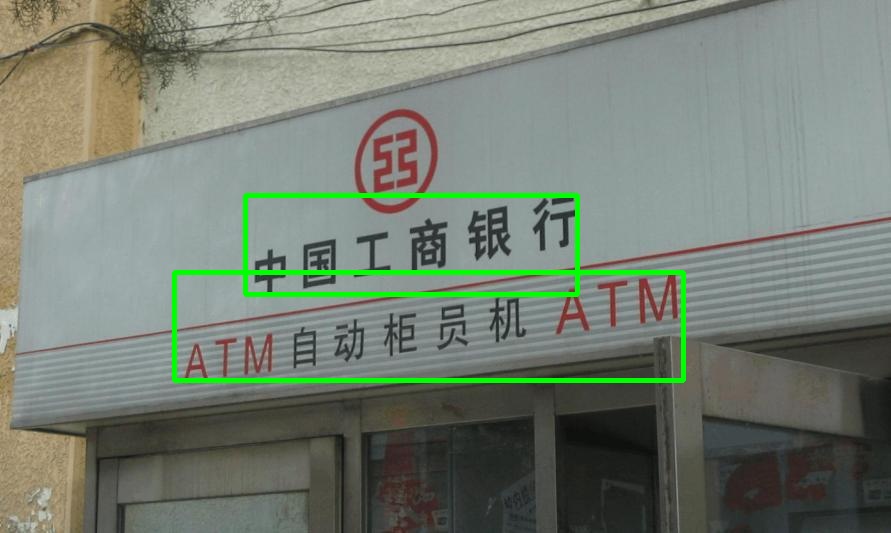} 
			\includegraphics[width=0.18\linewidth, height=\factorz\textheight]{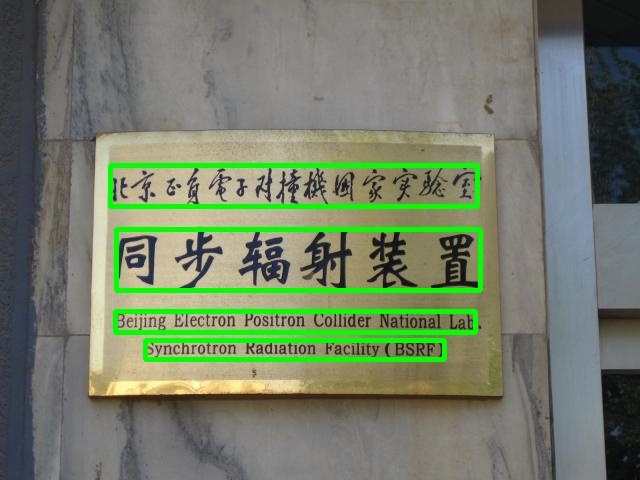}
			 \\ \vspace{1.5mm} 
			\includegraphics[width=0.18\linewidth, height=\factorz\textheight]{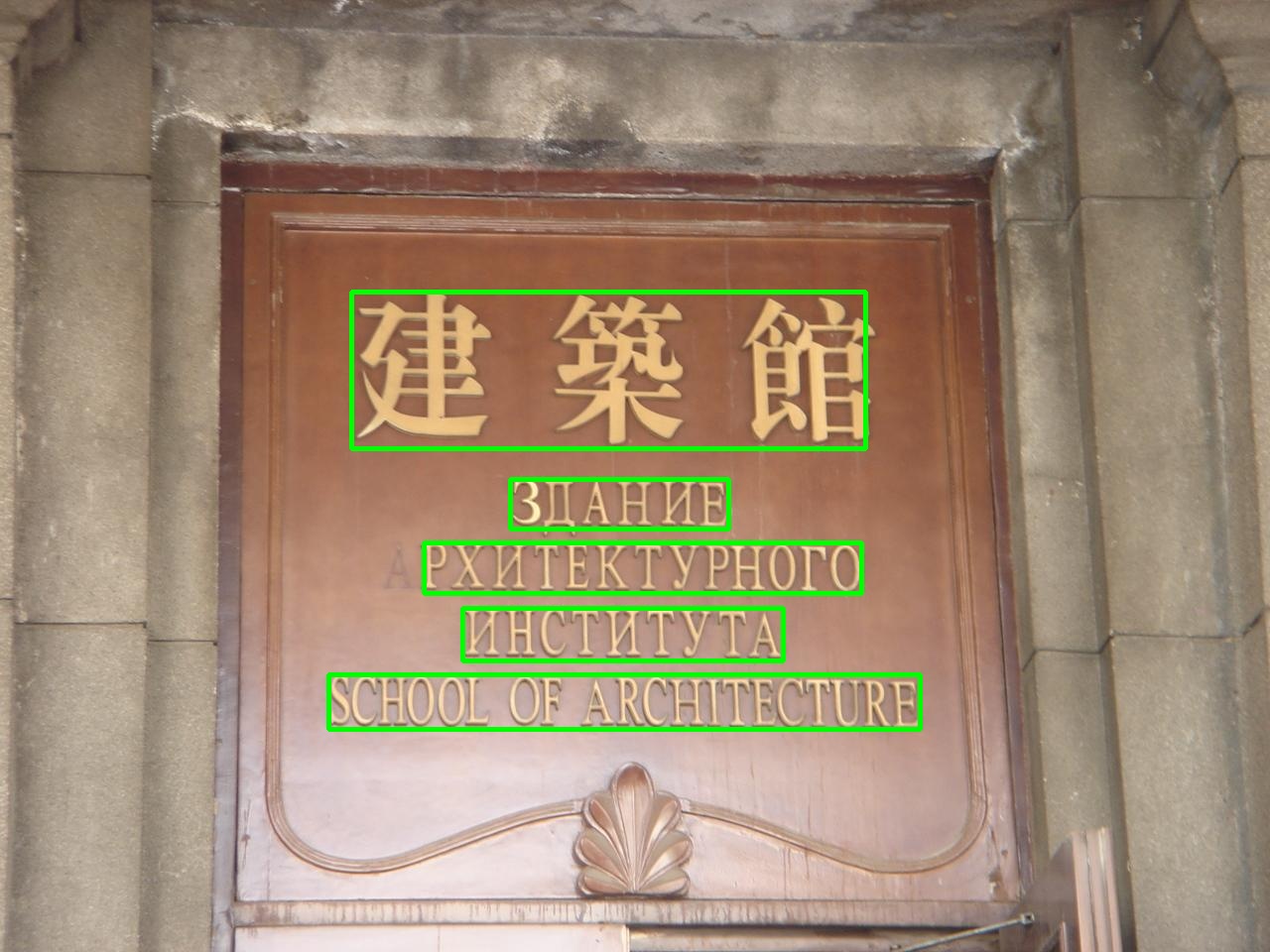}
			\includegraphics[width=0.18\linewidth, height=\factorz\textheight]{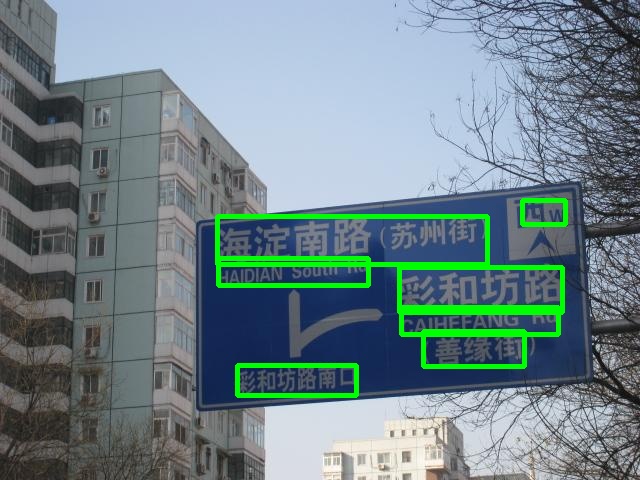} 
			\includegraphics[width=0.18\linewidth, height=\factorz\textheight]{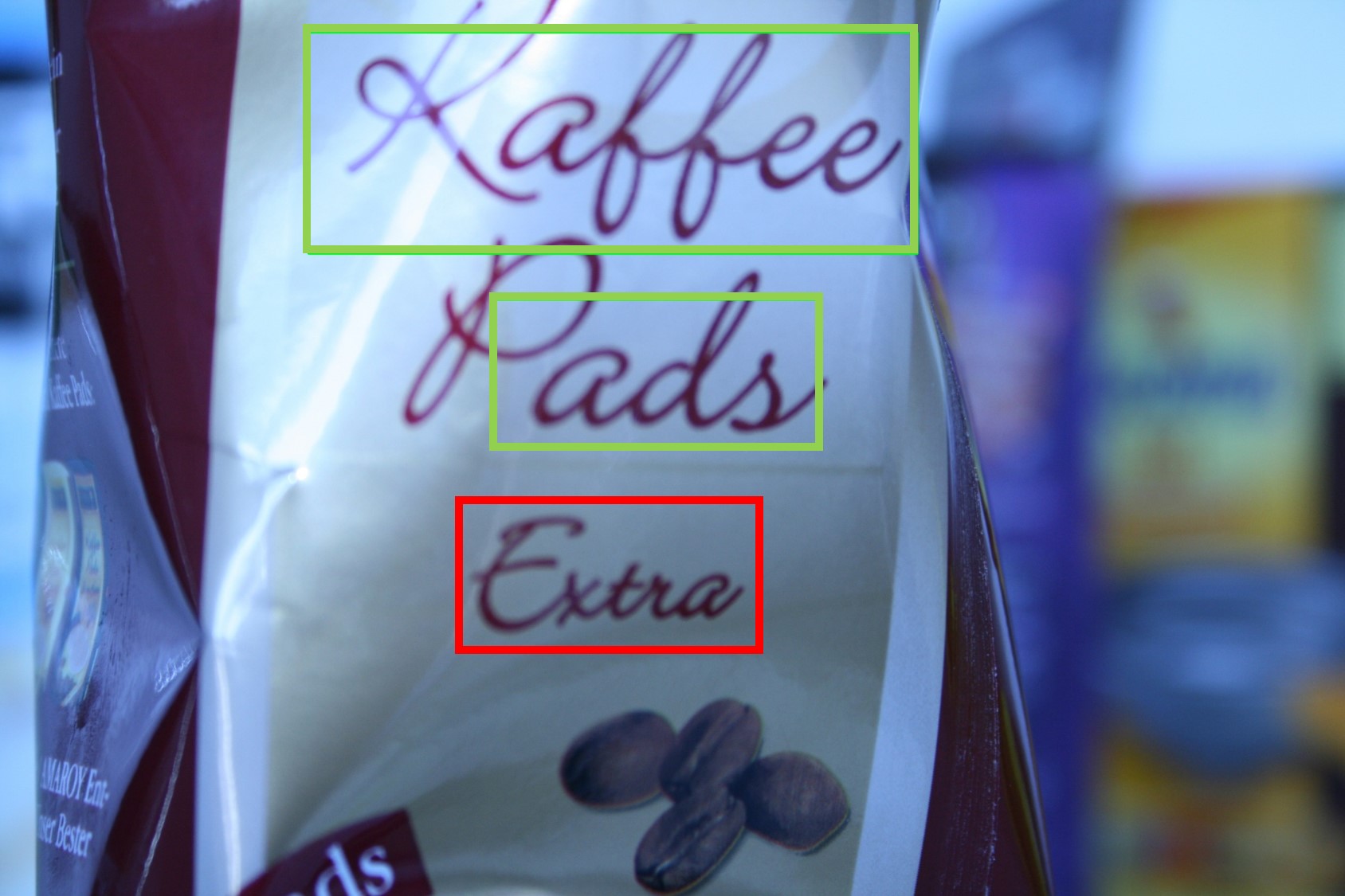}
			\includegraphics[width=0.18\linewidth, height=\factorz\textheight]{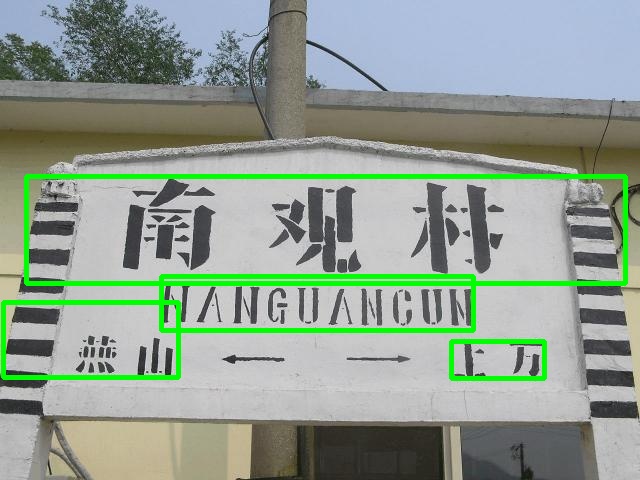}
			\includegraphics[width=0.18\linewidth, height=\factorz\textheight]{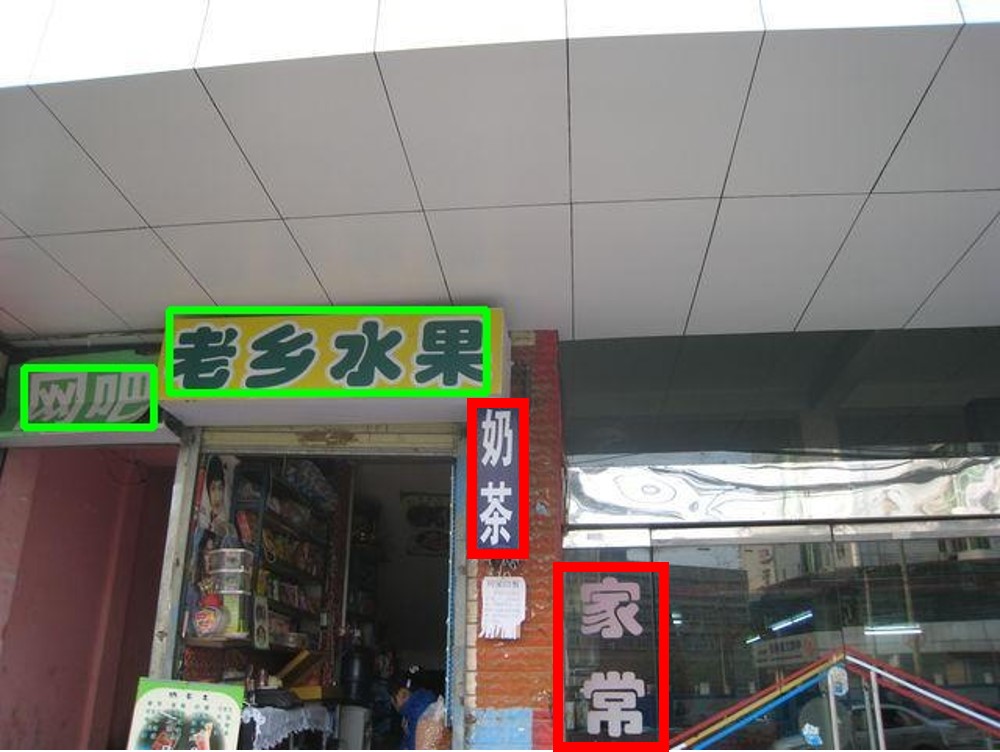}
	\caption{{Successful scene text detection examples on the ICDAR datasets (first three samples) and the multilingual dataset (ensuing four samples). Representative failure cases (last three samples) are illustrated, most of which suffer from the typical image degradation such as complex background, rare fonts, etc. Miss-detections are labeled by red bounding boxes.}}
	\label{Fig:show_result_exp}
\end{figure*}

To verify that the good performance is attributed to the min-cost flow model instead of the CNN classifier, a baseline scene text detection system is implemented for comparison. The system consists of the four sequential components, i.e, character candidate detection (same as the proposed method), character candidate elimination, text line extraction and text line verification. The text candidate and text line elimination is done by thresholding the CNN scores. While the text line extraction follows the methods in \cite{epshtein2010detecting, huang2014robust, sun2014robust}  which iteratively merge two text lines with similar geometric and heuristic properties. The result of the baseline system is shown in Table {\ref{tab:icdar13_res}} which decreases more than 7\%  in F-score. This comparison justifies that the min-cost flow model contributes much more to the good performance than the CNN classifier.

%

Figure \ref{Fig:show_result_exp} shows several sample results from the three public datasets. As we observe, the proposed technique works well on shaded texts, uncommon fonts, low contrast texts. {Besides, it can
detect both English and Chinese present in one image with the same character detector, as illustrated in Figure \ref{Fig:show_result_exp}. In addition, those characters with multiple isolated components are easily addressed in our framework while it may be quite complicated for CCs based methods to group those components into a character. Those facts demonstrate the superiority of the proposed method being utilized as a general text detection framework regardless of the language scripts. } On the other hand, the proposed method could miss some true text or detect non-text objects falsely under certain challenging conditions such as rare handwriting font, text similar patterns, vertical texts, etc.

\subsection{Discussion}

The good performance of our proposed technique is largely due to the min-cost flow network model which integrates multiple steps into a single process and accordingly solves the typical error accumulation problem. In particular, the min-cost flow network model incorporates the character level text confidence and the inter-character level spatial layout jointly which outperforms those approaches that exploit either character level confidence or inter-character level spatial layout alone. In addition, the good performance is also partially due to the cascade boosting model which gives a high recall of character candidates as well as the CNN employed in the min-cost flow network which provides reliable character confidence.



The proposed text detection system runs on a 64-bit Linux desktop with a 2.00GHz processor. On ICDAR2011 dataset, the processing time of MSER based methods are 0.43s and 1.8s per image, respectively as reported in {\cite{DBLP:dblp_journals/pami/YinYHH14}} and {\cite{neumann2012real}}. In comparison, the average processing time of the proposed method is 1.4s per image. On the multilingual dataset, our system is much faster than the hybrid method {\cite{pan2011hybrid}} and is comparable with the MSER based method {\cite{DBLP:dblp_journals/pami/YinYHH14}} as shown in Table {\ref{tab:pan_res}}. Though the sliding window scheme is adopted for the text candidate detection,  the proposed technique is quite fast because of the accelerating strategy in the character candidate detection step and the efficient min-cost flow network solution. 




\section{Conclusion}

In this paper, a novel text detection system, Text Flow, is proposed. The system consists of two steps including character candidate detection handled by cascade boosting and text line extraction solved by a min-cost flow network. The proposed system can capture the whole character instead of isolated character strokes and the processing time is comparable with the CCs based techniques. To handle the typical error accumulation problem, a flow network model is designed to integrate the three sequential steps into a single process which is solved by a min-cost flow technique. Experiments on the ICDAR2011 dataset, ICDAR2013 dataset and the multilingual dataset show that the proposed technique outperforms the state-of-the-art techniques greatly. Besides, the proposed system is superior in detecting texts in other non-Latin languages with a competitive speed compared to CCs based \mbox{methods}. 

{\small
\bibliographystyle{ieee}
\bibliography{egbib}
}

\end{document}